%% file: main.tex
\documentclass[10pt,twocolumn,letterpaper]{article}

\usepackage{wacv}              

\usepackage{graphicx}
\usepackage{amsmath}
\usepackage{amssymb}
\usepackage{booktabs}
\usepackage{tikz}

\input{math_commands.tex}

\usepackage{graphbox}
\usepackage{multirow}
\usepackage{makecell}

\usepackage[accsupp]{axessibility}  

\usepackage[pagebackref,breaklinks,colorlinks]{hyperref}
\usepackage{subcaption}

\usepackage[capitalize]{cleveref}
\crefname{section}{Sec.}{Secs.}
\Crefname{section}{Section}{Sections}
\Crefname{table}{Table}{Tables}
\crefname{table}{Tab.}{Tabs.}

\begin{document}

\title{AH-OCDA: Amplitude-based Curriculum Learning and Hopfield Segmentation Model for Open Compound Domain Adaptation}

\author{\textbf{Jaehyun Choi\thanks{Equal contribution.}\hspace{1.3cm} Junwon Ko\footnotemark[1]\hspace{1.3cm} Dong-Jae Lee\footnotemark[1]\hspace{1.3cm} Junmo Kim}\vspace{0.3cm} \\
School of Electrical Engineering, KAIST, South Korea\\
{\tt\small \{chlwogus, kojunewon, jhtwosun, junmo.kim\}@kaist.ac.kr}}


\maketitle

\input{sections/0_abstract}
\input{sections/1_intro}
\input{sections/2_related_works}
\input{sections/3_method}
\input{sections/4_experiment}
\input{sections/5_conclusion}
\input{sections/6_ack}

\bibliographystyle{splncs04}
\bibliography{main}

\clearpage
\setcounter{section}{0}
\renewcommand{\thesection}{\Alph{section}}
\input{supplementary_sections/0_dataset}
\input{supplementary_sections/1_hopfield_high-level}
\input{supplementary_sections/2_qualitative_result}
\input{supplementary_sections/3_limitations}

\begin{figure*}
    \centering
    \includegraphics[width=1.02\textwidth]{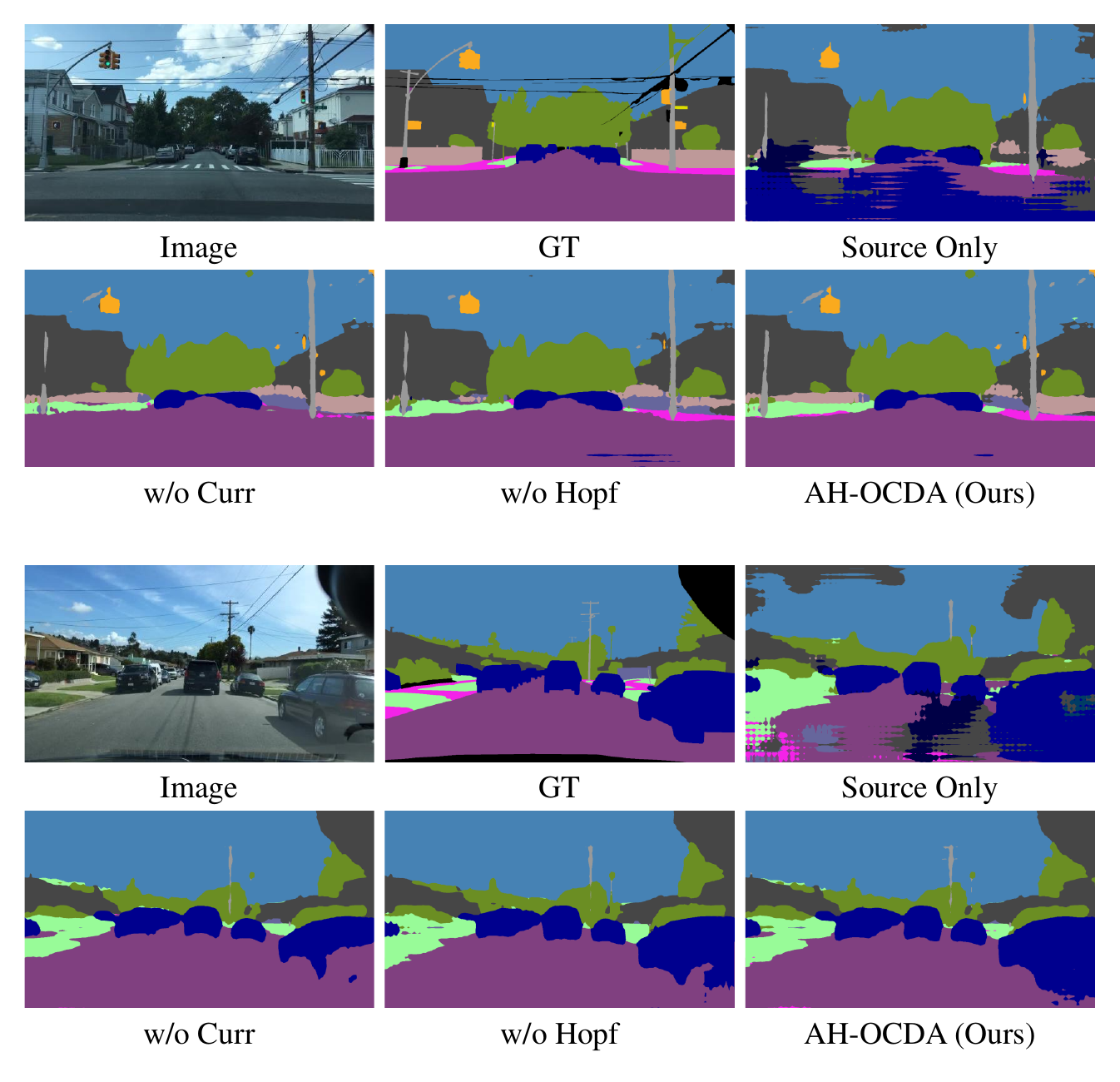}
    \caption{Qualitative analysis on `Cloudy' domain of GTA5 $\to$ C-Driving}
    \label{fig:sup_cloudy}
\end{figure*}

\begin{figure*}
    \centering
    \includegraphics[width=1.02\textwidth]{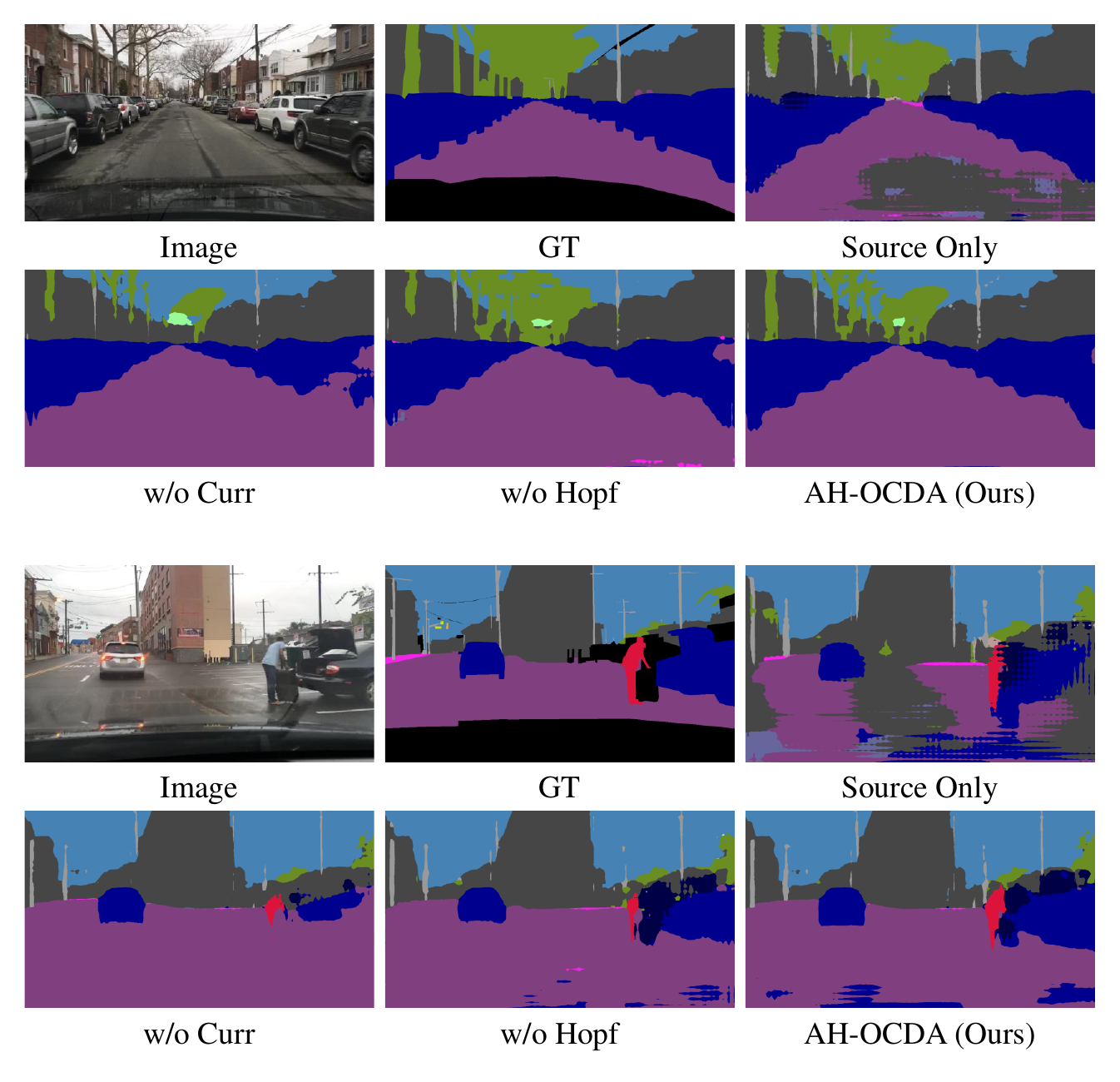}
    \caption{Qualitative analysis on `Rainy' domain of GTA5 $\to$ C-Driving}
    \label{fig:sup_rainy}
\end{figure*}

\begin{figure*}
    \centering
    \includegraphics[width=1.02\textwidth]{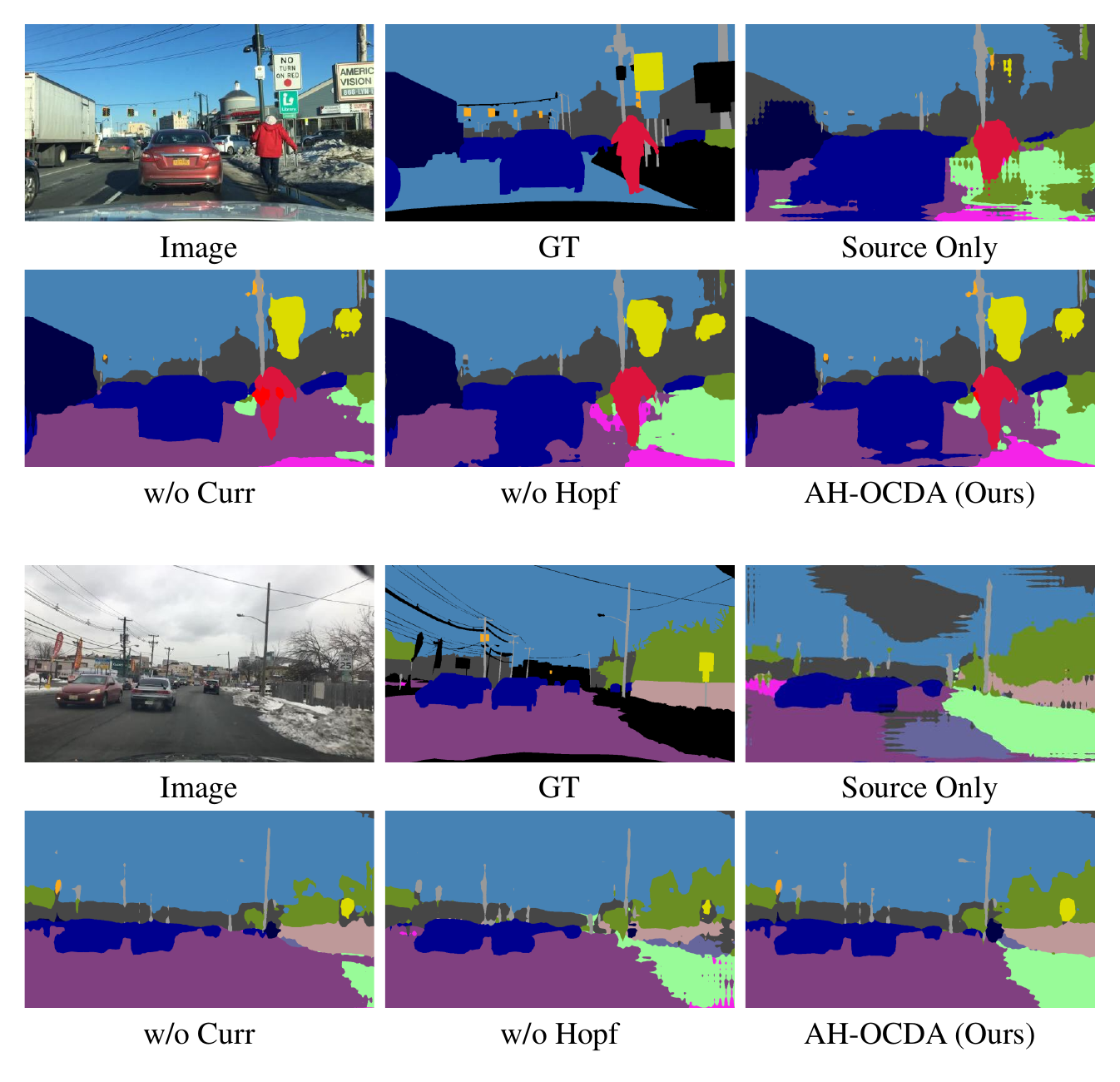}
    \caption{Qualitative analysis on `Snowy' domain of GTA5 $\to$ C-Driving}
    \label{fig:sup_snowy}
\end{figure*}

\begin{figure*}
    \centering
    \includegraphics[width=1.02\textwidth]{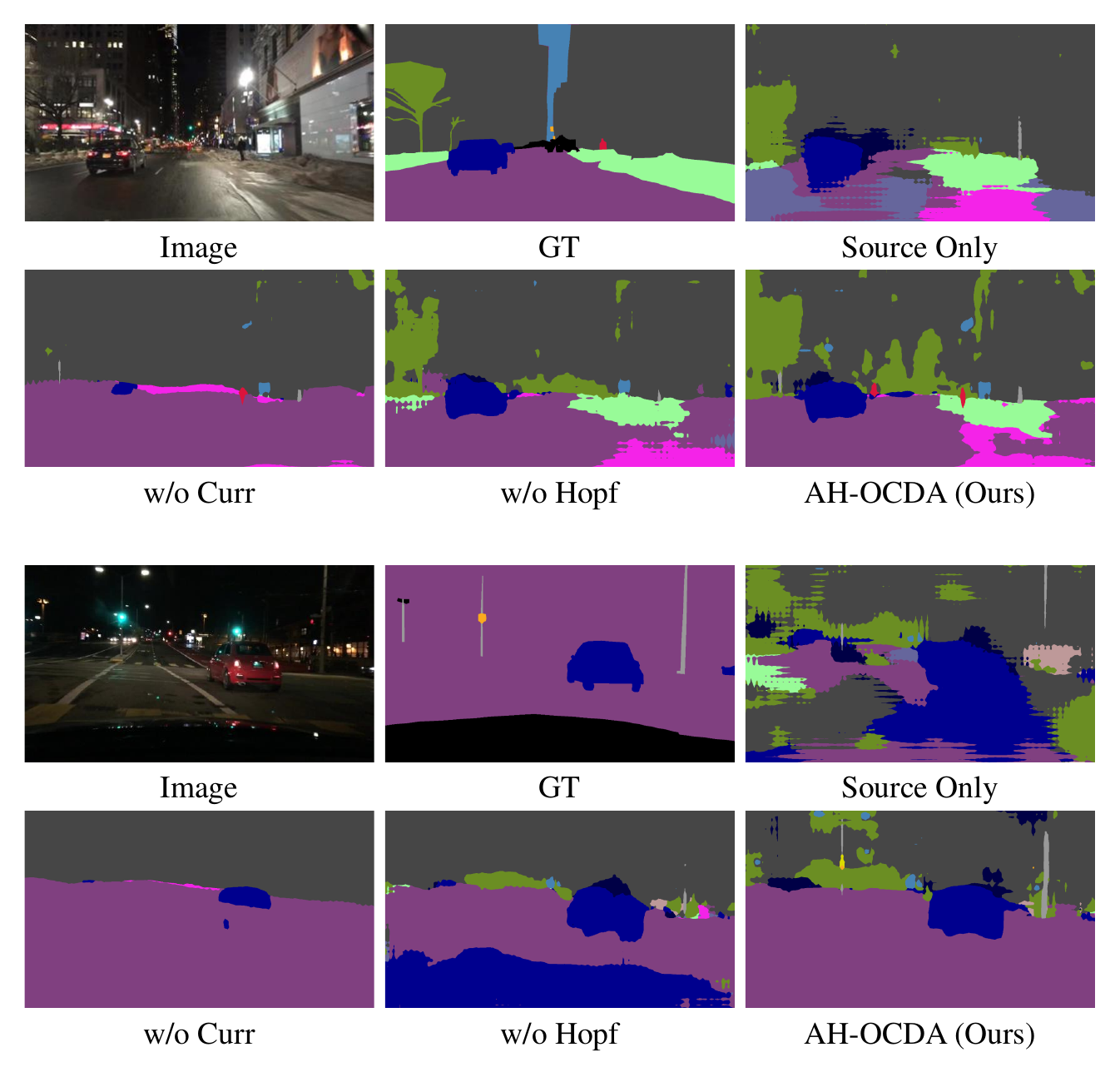}
    \caption{Qualitative analysis on `Night' domain of GTA5 $\to$ C-Driving}
    \label{fig:sup_night}
\end{figure*}

\begin{figure*}
    \centering
    \includegraphics[width=1.02\textwidth]{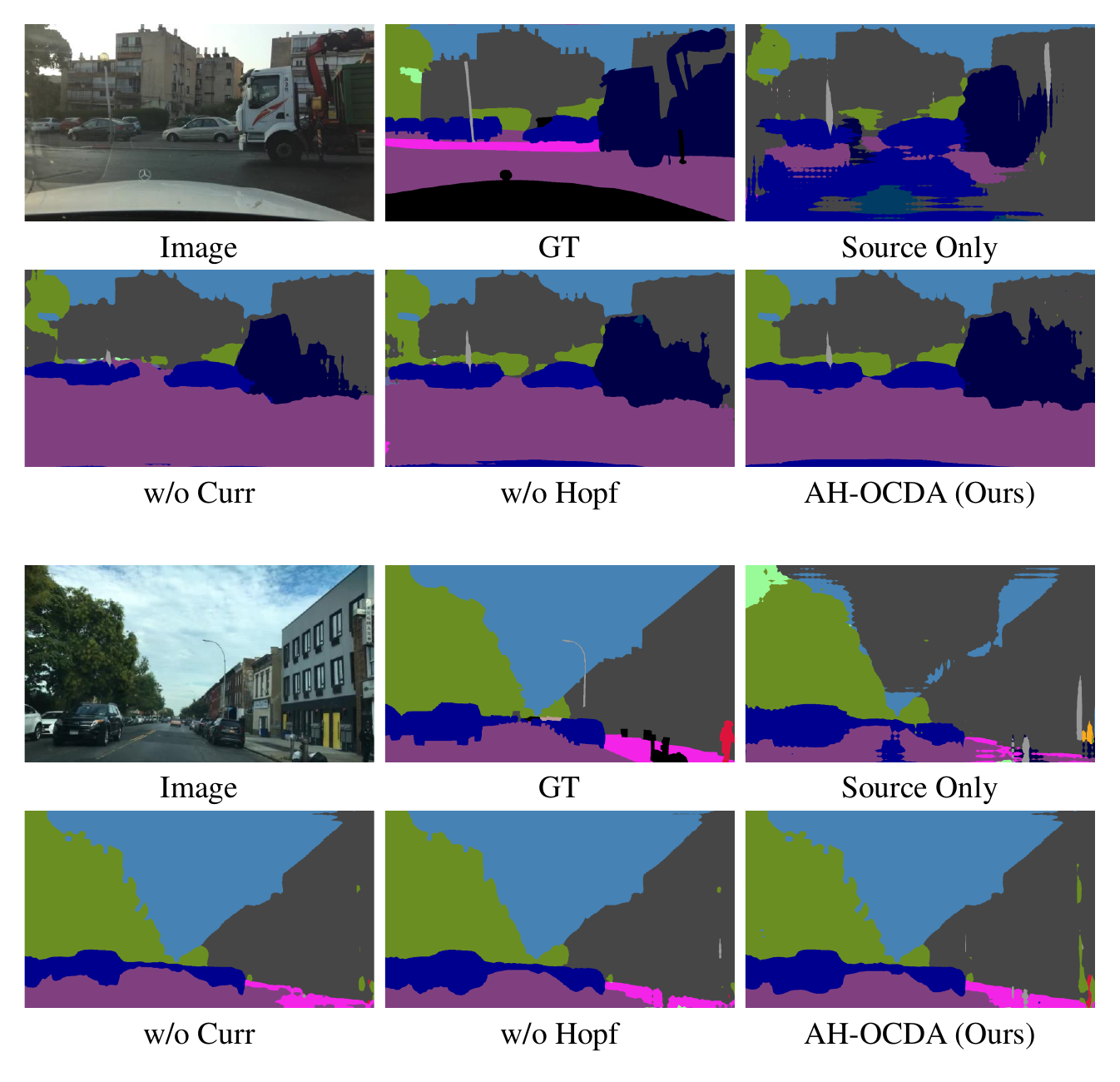}
    \caption{Qualitative analysis on `Overcast' domain of GTA5 $\to$ C-Driving}
    \label{fig:sup_overcast}
\end{figure*}

\end{document}

%% file: math_commands.tex
\usepackage{amsmath,amsfonts,bm}

\def\eqref#1{equation~\ref{#1}}

\def\1{\bm{1}}

\DeclareMathAlphabet{\mathsfit}{\encodingdefault}{\sfdefault}{m}{sl}
\SetMathAlphabet{\mathsfit}{bold}{\encodingdefault}{\sfdefault}{bx}{n}



%% file: sections/0_abstract.tex
\begin{abstract}
Open compound domain adaptation (OCDA) is a practical domain adaptation problem that consists of a source domain, target compound domain, and unseen open domain.
In this problem, the absence of domain labels and pixel-level segmentation labels for both compound and open domains poses challenges to the direct application of existing domain adaptation and generalization methods.
To address this issue, we propose Amplitude-based curriculum learning and a Hopfield segmentation model for Open Compound Domain Adaptation (AH-OCDA).
Our method comprises two complementary components: 1) amplitude-based curriculum learning and 2) Hopfield segmentation model.
Without prior knowledge of target domains within the compound domains, amplitude-based curriculum learning gradually induces the semantic segmentation model to adapt from the near-source compound domain to the far-source compound domain by ranking unlabeled compound domain images through Fast Fourier Transform (FFT).
Additionally, the Hopfield segmentation model maps segmentation feature distributions from arbitrary domains to the feature distributions of the source domain.
AH-OCDA achieves state-of-the-art performance on two OCDA benchmarks and extended open domains, demonstrating its adaptability to continuously changing compound domains and unseen open domains.
\end{abstract}

%% file: sections/1_intro.tex
\section{Introduction} \label{sec_intro}

Semantic segmentation has been explored in various areas, including medical image processing, autonomous driving, and robot scene understanding.
However, the need for meticulous pixel-level labels in training presents a significant challenge for direct model deployment.
To mitigate the burden of acquiring such labels, domain adaptation and domain generalization have been investigated.
These tasks primarily focus on minimizing the distribution difference between the source and target domains, also known as the domain gap.
Although they have demonstrated remarkable performance, each approach has certain limitations: domain adaptation experiences performance degradation in novel target domains, while domain generalization requires obtaining pixel-level labels across multiple domains.
This challenge has driven the development of approaches such as multi-target domain adaptation, single-source domain generalization, and recently, open compound domain adaptation (OCDA).

OCDA, first proposed by \textit{Liu et al.}~\cite{liu2020ocda}, leverages the advantages of aforementioned tasks and adapts to multiple unlabeled target domains (a compound domain) and novel domains (an open domain) with a single labeled source domain.
Specifically, OCDA treats the compound domain as one continuous spectrum of potential target domains, enhancing its practicality even when the domain labels within the real-world target domains are vague.
Additionally, by addressing both the compound and open domains simultaneously, OCDA further enhances its practicality, aligning with the ideal scenario in which semantic segmentation models are trained on labeled synthetic source domains but deployed in environments where domains are unknown.
Previous works~\cite{liu2020ocda, park2020dha, gong2021csfu} solve this complicated problem by dividing the compound domain into multiple discrete domains, narrowing the task into several unsupervised domain adaptation problems.
While these approaches have demonstrated promising performance, they are limited by training separate segmentation models, requiring sophisticated hyperparameter tuning for each domain and model while also not fully accounting for the open domain.

To overcome the limitations, we propose a novel method, amplitude-based curriculum learning with the Hopfield segmentation model for OCDA (AH-OCDA).
AH-OCDA integrates two key components: amplitude-based curriculum learning and the Hopfield segmentation model.
The amplitude-based curriculum learning is a ranking-based approach that leverages the frequency spectrum of each image to treat the compound domain as a continuous spectrum of target domains.
Moreover, the Hopfield segmentation model maps distorted features to source features using a modern continuous Hopfield network, aiming to overcome the challenge of unseen open domains.
It is noteworthy that these two components are orthogonal, complementing each other to achieve better performance in OCDA.

Amplitude-based curriculum learning leverages domain-specific information of the image, drawing upon the insight that amplitude in the low-frequency spectrum contains the domain-specific information~\cite{yang2020fda, fftdg}.
Specifically, we extract the low-frequency component from the amplitude obtained by the Fast Fourier Transform (FFT) of the target image and compute mean squared error (MSE) with the mean amplitude of source domain images.
We continuously adapt the model trained in the source domain to the near-source target domain, while utilizing the previously adapted target domain as \textit{fake source} with pseudo labels.
The proposed curriculum learning only requires knowledge of the individual gap between target images and the source domain, maintaining robustness against dynamic domain shifts.
Meanwhile, the Hopfield segmentation model maps segmentation feature distributions from arbitrary domains to the feature distributions of the source domain by learning and retrieving feature patterns.
Modern continuous Hopfield network \cite{hopfieldall, uhn} learns relational feature mappings, enabling it to restore corrupted or noisy patterns to their original pattern by leveraging associative memory.
Building on this capability, the Hopfield segmentation model implicitly stores representative information within its parameters, including intra-class and inter-class information.
This allows the network to accurately predict distorted feature vectors from compound or open domains previously encountered.
Unlike approaches that rely on manually modeling feature distribution statistics, such as memorizing the class centroids or statistical information, our Hopfield segmentation model requires only the source domain to learn raw patterns and map the target domain features to their corresponding source domain features.

While the Hopfield segmentation model is capable of mapping arbitrary domain features to source domain features, it cannot guarantee the successful mapping of features from significantly distant domains to those in the known domain.
At this point, amplitude-based curriculum learning trains the Hopfield segmentation model gradually, starting from the near-source target domain to the far-source target domain, enabling the model to map target domain features to the source domain features.
By gradually adapting to target domains through adversarial learning and using \textit{fake source}, the network can fully leverage the memory retrieval process to progressively enlarge the boundary of the known domain.
Conversely, the Hopfield network within the segmentation model enables the model to predict domains beyond the known boundary established by curriculum learning by restoring distorted features to the corresponding source features.
Our experiments confirmed that these components function orthogonally, significantly enhancing the model's performance when combined.
Furthermore, we validated effectiveness of AH-OCDA on two OCDA benchmarks and additional extended open domains, where we achieved state-of-the-art performance.

%% file: sections/2_related_works.tex
\section{Related works} \label{sec_related}
\subsection{Domain Adaptation and Generalization} \label{rel_da}
Domain adaptation (DA) is a key technique in machine learning where a model trained on one domain is adapted to perform well on a different, but related domain.
Unsupervised domain adaptation (UDA) is a specific approach in which the model is adapted to an unlabeled target domain.
They have advanced to scenarios with single target domain \cite{saenko2010DAsingle1, venkateswara2017DAsingle2, peng2019DAsingle3} or clearly distinct multi-target domains \cite{gong2013DAmulti1, gholami2020DAmulti2, yu2018multi3} and also to scenario with multiple source domain~\cite{multisourceda1, multisourceda2}.
In parallel, the concept of domain generalization (DG) aims to train a model on multiple source domains to be generalized on unseen target domains \cite{dou2019dg1, huang2020dg2, blanchard2021dg3, fftdg, dg1} or train a model on single source domains to be generalized on unseen target domains~\cite{singlesourcedg1, singlesourcedg2}.
However, these methods may not always capture the full complexity of real-world scenarios.
In practical applications, data collection often results in mixed, continuous, and unseen domains, highlighting the need for new problem settings that better reflect these diverse and dynamic real-world conditions.

\subsection{Open Compound Domain Adaptation} \label{rel_ocda}
Open compound domain adaptation (OCDA) \cite{liu2020ocda} suggested a problem setting that has a labeled source domain and a compound of multiple unlabeled target domains for training phase and unseen target domains for testing phase.
They proposed a curriculum learning framework for semantic segmentation tasks, which schedules the curriculum based on the prediction confidence of a model that is only trained on the source domain.
On the other hand, the following studies did not follow OCDA's approach but proposed their own unique methods.
DHA \cite{park2020dha} and MOCDA \cite{gong2021csfu} assigned pseudo domain labels with K-means clustering to break down OCDA into multiple UDA settings.
Specifically, DHA \cite{park2020dha} utilizes K-means clustering on the convolutional feature statistics, which are mean and standard deviations.
They stylize the source domain into each target latent domain to reduce the domain gap and adversarially align a shared segmentation model employing individual discriminators for each pair of stylized source and target latent domains.
MOCDA \cite{gong2021csfu} utilizes K-means clustering on style code, which is low-dimensional space generated by an additional encoder.
They adopt a model-agnostic meta-learning approach, where a hyper-network combines domain-specific segmentation models and discriminators to consider the continuous compound target domain.
However, K-means clustering may not be a suitable approach for open compound domain adaptation, as it may not effectively represent continuous target domains and accommodate expandable unseen domains.
To address this limitation, we explore a more dynamic approach.

\subsection{Fourier Transform for Domain Adaptation} \label{rel_fft}
Several works have discussed how the Fourier transform represents the style information of an image in its magnitude spectrum and have applied this concept to domain adaptation \cite{da_styletransfer1, yang2020fda, huang2021fft2, da_styletransfer2, li2023fftstyle, zhong2023bifreqalign, da_styletransfer3}.
In this perspective, FDA \cite{yang2020fda} proposed a Fast Fourier Transform (FFT)-based style transfer that reduces the style gap between two domains by replacing the low-frequency spectrum of the images.
Building on FDA, some methods have further developed FFT-based style transfer for DA by incorporating the ordering of target data during the training phase, similar to approaches used in continual learning~\cite{da_styletransfer2, da_styletransfer3}.
BAT \cite{zhong2023bifreqalign} further applied FFT-based style transfer for multi-target domain adaptation instead of network-based style transfer.
To cope better with OCDA task, AST-OCDA \cite{kundu2022ampspectrans} suggested applying the FT-based style transfer on the feature space of the segmentation model instead of the input RGB image.
For single-domain or discrete multiple-domain adaptation, applying Fourier-based style information at a particular feature level has proven effective.
However, in OCDA, the direct application of Fourier-based style transfer is less feasible for adapting the source domain to diverse, continuous target domains, and unseen domains.
To overcome this, we considered the flexible utilization of domain-specific information in OCDA scenarios.

\subsection{Associative Memory} \label{rel_hn}
Associative memory systems function by retrieving data not through specific addresses but via queries that resemble the type of data stored within them \cite{hopfield2007hopfield}. 
When a query is made, the system identifies and returns a data point that most closely matches the query. For example, presenting an image as a query would yield other images that are deemed similar.
This retrieval process is often compared to how the human brain operates, where memories can be recalled with partial cues, such as remembering a song from a few notes \cite{uhn}. 
Two classical and influential models are the Hopfield network (HN) and the sparse distributed memory (SDM).
More recently, they have been generalized to the modern continuous Hopfield network (MCHN), which substantially improves performance and can handle continuous inputs.
Specifically, MCHN uses the energy function $E = q^\top q + \log\sum\exp(Wq)$, which can be minimized using the convex-concave method, giving the update rule $z = W^\top \sigma(Wq)$, where $\sigma(\cdot), W, q$ represent the softmax function, the memory matrix and the query vector, respectively.
This is similar to the attention operation $z = V\sigma(Wq) $ with key matrix K and value matrix V, which is widely used in recent neural networks \cite{xie2021segformer, daformer, lee2023lightweight}, where we can associate Q = q, K = W, and V = W \cite{hopfieldall}.

%% file: sections/3_method.tex
\begin{figure*}[h]
    \centering
    \includegraphics[width=1\linewidth]{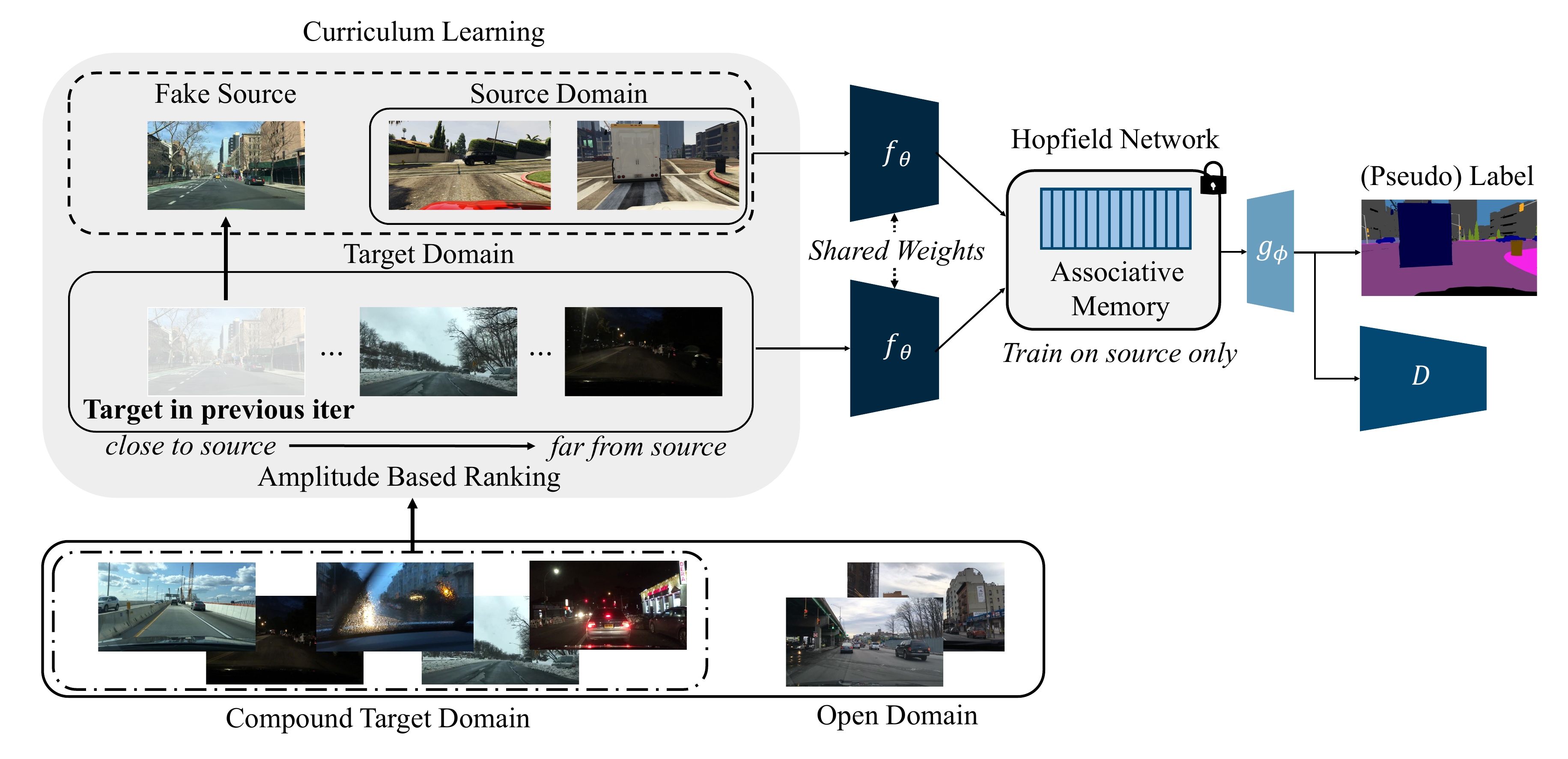}
    \caption{Overview of the proposed pipeline. For the amplitude-based curriculum learning, we extract the amplitudes with the Fast Fourier Transform and measure the distance between each target image and the source domain images. The Hopfield segmentation model is frozen when training.}
    \label{fig:pipeline}
\end{figure*}

\section{Method} \label{sec_method}
Our approach for OCDA in semantic segmentation consists of two components: 1) amplitude-based curriculum learning and 2) the Hopfield segmentation model.
Figure \ref{fig:pipeline} illustrates our whole pipeline.

\subsection{Preliminaries}
Let spatial domain space $\mathcal{X} \subset \mathbb{R}^{H \times W \times 3}$ and frequency domain space $\mathcal{A} \subset \mathbb{C}^{H \times W \times 3}$, where H and W are height and width of the input image, respectively.
Through FFT $\mathcal{F} : \mathcal{X} \to \mathcal{A}$, we transform the image $x \in \mathcal{X}$ to the spectrum $\mathcal{F}(x)$.
Then, we obtain amplitude $|\mathcal{F}(x)| \in \mathbb{R}^{H \times W \times 3}$ and phase $\texttt{Arg}(\mathcal{F}(x)) \in \mathbb{R}^{H \times W \times 3}$.
The semantic segmentation model $Seg$ consists of an encoder $f$ and classifier $g$, such that $Seg: x \mapsto y = g(f(x))$, where $y \in \mathbb{R}^{H \times W}$ represents semantic segmentation prediction.

Following OCDA, the labeled source domain consists of synthetic RGB images $x_s$ and corresponding semantic segmentation label $y_s$, such that
\begin{equation}
\mathcal{S} = \left\{(x_s^i, y_s^i) \mid x_s^i \in \mathcal{X}, y_s^i\right\}, i = 1, \ldots, N_s
\end{equation}
where $N_s$ is the number of images in the labeled source domain.
The unlabeled compound domain consists of real RGB input images $x_c$, such that
\begin{equation}
\mathcal{C} = \left\{x_c^i \mid x_c^i \in \mathcal{X}\right\}, i = 1, \ldots, N_c
\end{equation}
where $N_c$ is the number of images in the unlabeled compound domain.
The unseen domain consists of real RGB images $x_u$, such that 
\begin{equation}
\mathcal{U} = \left\{x_u^i \mid x_u^i \in \mathcal{X}\right\}, i = 1, \ldots, N_u
\end{equation}
where $N_u$ is the number of images in the unseen domain.

\subsection{Amplitude-based Curriculum Learning} \label{met_curri}
The amplitude-based curriculum learning is a ranking-based approach that only requires knowledge of the individual gap between the target image and the source domain images.
First, we transform the image $x_i$ to the spectrum through the FFT.
Then we shift values making the center of the spectrum to represent the low-frequency region.
Finally, we obtain the domain-specific information by utilizing only the amplitude from the spectrum and cropping the central region, a region that is known to contain the domain-specific information.
\begin{equation}
F_\beta^i = \textit{Crop}_{\beta}\left\{\left|\mathcal{F}(x_i)\right|\right\} \in \mathbb{R}^{\beta H \times \beta W \times 3}
\end{equation}
where $\beta$ is the cropping ratio of height and width in the spectrum.
The average domain information of the source domain is defined as below.
\begin{equation}
\mathcal{M}_s = \frac{1}{N_s}\sum_{x_s \in \mathcal{S}}\textit{Crop}_{\beta}\left\{\left|\mathcal{F}(x_s)\right|\right\}
\end{equation}
Once $\mathcal{M}_s$ is obtained, we can define the domain distance ${\delta}$ of the compound image $x_c$ like below.
\begin{equation}
\delta_c = (\mathcal{M}_s - F_\beta^c)^2
\end{equation}
We rank $x_c\in\mathcal{C}$ to define curriculum $\textit{G}$ based on their $\delta_c$, arranging them in ascending order.
\begin{equation}
\textit{G} = \{ x_{c} \mid x_{c} \text{ is sorted in ascending order for } \delta_c \}
\end{equation}
For the adversarial alignment, we divide $\textit{G}$ into $K$ groups where $K$ is a factor of size of $\textit{G}$.
\begin{equation}
\textit{G} = \bigcup_{j=1}^{K} \mathcal{T}_j
\end{equation}
where each $\mathcal{T}_j$ contains $N_c/K$ number of $x_c$.
The learning process gradually includes $\mathcal{T}_j$ to perform near-source to far-source curriculum learning.

The segmentation model gets $x_s$ and $x_c$ from the curriculum and predicts the softmax output $\hat{y}_s$ and $\hat{y}_c$.
Then, the discriminator takes $\hat{y}_s$ and $\hat{y}_c$ to differentiate whether the input originates from the source or the target domain.
In this process, adversarial learning induces the segmentation model to be unable to distinguish between source and target domain images.
However, na\"ively applying curriculum learning has a limitation as the previously leveraged target images $\mathcal{T}_{j-1}$ will no longer be utilized for training and thus may result in catastrophic forgetting.
For example, when the pipeline is on $\mathcal{T}_j$, the previous $x_c$ in $\mathcal{T}_{j-1}$ is no longer being used for training.
In this case, the performance on $x_c$ will get deteriorated reducing overall domain adaptation performance on $\mathcal{C}$.
To address this issue, our framework assigns half of previously adapted target domain images as a fake source with the pseudo-label from prediction along with source domain images as $\hat{\mathcal{S}}_j$ where $j$ is the step in which curriculum learning is currently in.
\begin{equation}
\hat{\mathcal{S}}_j = \mathcal{S} \cup \mathcal{P}_j
\end{equation}
where $\mathcal{P}_j = \{x_c^1, x_c^2, \dots, x_c^{\lfloor\frac{|G|\times (j-1)}{2K}\rfloor}\}$.
The fake-source approach ensures that the framework continuously adapts with the combined set of $\mathcal{S}$ and the fraction of $x_c$.

\subsection{Hopfield Segmentation Model} \label{met_hopfield}
In contrast to traditional domain generalization scenarios, where multiple labeled domains are typically available, OCDA contains labels only in the source domain, while both the compound and open domains do not have segmentation and domain labels during training.
This challenge results in suboptimal performance of previous methods on target domains that significantly deviate from the source domain, as well as on unseen open domains.
To address this challenge, we devised the Hopfield segmentation model to empower the semantic segmentation pipeline with the capability to align features in the target domain with their corresponding source domain, even on the novel unseen domains.
Intuitively, adaptation for compound and open domains can be easily solved by mapping the arbitrary target features to the source feature distribution.
The Hopfield network in the segmentation model correct distorted feature patterns or noisy inputs from unfamiliar domains by retrieving stored source patterns.
In other words, given the encoder features $z$, our objective is to learn the mapping function $H(\cdot)$ that maps the arbitrary feature to the source feature distribution.

Mathematically, the associative memory is a function $g(\cdot): \mathbb{R}^{C_i} \rightarrow \mathbb{R}^{C_o}$ mapping a vector in an input space of $C_i$ to a vector in an output space of dimension $C_o$, with two additional inputs of a memory matrix $M \in \mathbb{R}^{M_N \times C_i}$ consisting of a set of $M_N$ stored patterns, and projection matrices $P \in \mathbb{R}^{C_i \times C_o}$, consisting of a potentially different set of stored patterns with dimension $C_o$ for hetero-association.
Similarly, our Hopfield network consists of a memory matrix $M \in \mathbb{R}^{M_N \times C_l}$ and query, key, and value projection matrices $W_q, W_k \in \mathbb{R}^{C_l \times C_s}$ and $ W_v \in \mathbb{R}^{C_s \times C_s}$, where $M_N, C_l, C_s$ represents the memory size, dimension of the feature vector, and dimension of the projected feature vector for similarity measurement, respectively.

We first measure the similarity of the feature embedding $z$ and each raw stored pattern vector $m_i \in {M}_{N}$:
\begin{equation}
\text{sim}(z, m_i) = \frac{\text{exp}(\tau \cdot W_qz \cdot (W_km_i)^\top)}{\sum_{j \in M_N} 
 \text{exp}(\tau \cdot W_qz \cdot (W_km_j)^\top)}.
\end{equation}
where $\tau$ represents the scaling factor.
The role of each projection matrix is to project the raw state patterns $z$ and stored patterns $m_j$ to associative space and reduce computational complexity through dimension reduction ($C_s < C_l$).
Then, we project the $z$ to the stored pattern with the weighted sum of the similarity and the stored patterns:
\begin{equation}
\hat{z} = \sum_{j \in M_N} \text{sim}(z, m_j) W_v m_j.
\end{equation}
The projected feature $\hat{z}$ is fed to the classifier to predict the segmentation map and update its memory with the segmentation loss.

For training the Hopfield segmentation model, we pretrain the Hopfield network in the segmentation model for the source dataset.
Then, in the following adaptation phase, we freeze the associative memory.
Specifically, we freeze the memory weights with key and value projection weights.
Therefore, the memory network maps the feature from target domains to the trained source domain feature distribution.
The pretrained Hopfield network in the segmentation model is likely to be robust even on far-source target domains, as the amplitude-based curriculum learning gradually trains the entire segmentation model to obtain a well-structured target feature distribution.

\subsection{Loss Function} \label{met_loss}
Our framework utilizes the following loss to optimize the segmentation model  $Seg$ and the discriminator $D$:
\begin{equation}
\mathcal{L}(\hat{S}_j, \mathcal{T}_j) = \mathcal{L}_{ce}^{Seg}(\hat{S}_j) + \lambda_{\text{adv}}(\mathcal{L}_{\text{adv}}^{Seg}(\mathcal{T}_j) + \mathcal{L}_{\text{adv}}^\textit{D}(\hat{S}_j, \mathcal{T}_j)).
\end{equation}
This loss consists of two parts: $\mathcal{L}_{ce}^{\textit{S}}$ is the cross entropy loss that updates the segmentation model to make accurate predictions in the source domain and the \textit{fake-source}.
Formally, $\mathcal{L}_{ce}^{\textit{S}}$ is defined as
\begin{equation}
\mathcal{L}_{ce}^{Seg}(\hat{S}_j) = -\sum_{h,w} \sum_{cls} Y_{\hat{S}}^{(h, w, cls)} \log(Seg(\hat{S}_j)^{(h, w, cls)}),
\end{equation}
where $cls$ is the number of classes, and $Y_{\hat{S}}$ is the (pseudo) label of the (fake-) source image.
$\mathcal{L}_{\text{adv}}$ is applied separately to the segmentation model $\mathcal{L}_{\text{adv}}^{S}$ and the discriminator $\mathcal{L}_{\text{adv}}^D$, which are
\begin{equation}
\mathcal{L}_{\text{adv}}^{Seg}\left(\mathcal{T}_j\right)=-\sum_{h, w} \log \left(D\left(Seg\left(\mathcal{T}_j\right)\right)^{(h, w, 1)}\right),
\end{equation}
\begin{equation}
    \begin{array}{r}
    \mathcal{L}_{\text{adv}}^D(\mathcal{T}_j, \hat{S}_j)=-\sum_{h, w}\log \left(D(Seg\left(\mathcal{T}_j\right))^{(h, w, 0)}\right) \\
    + \log \left(D(Seg\left(\hat{S}_j\right))^{(h, w, 1)}\right).
    \end{array}
\end{equation}
$\mathcal{L}_{\text{adv}}$ induces the adversarial learning for the segmentation model and discriminator.
$\lambda_{\text{adv}}$ adjusts the importance between the two losses. 

%% file: sections/4_experiment.tex
\section{Experiments} \label{sec_experiments}
\subsection{Experimental settings} \label{sec:exp_settings}
For the labeled source domain we leverage GTA5 \cite{richter2016gta5} and SYNTHIA \cite{ros2016synthia} while BDD100K (C-Driving) \cite{yu2020bdd100k} is utilized for the unlabeled compound and open domains.
For extended open domain, we evaluate with Cityscapes \cite{cordts2016cityscapes} and KITTI \cite{geiger2013kitti}.
The details for each dataset can be found in the supplementary material.
For all experiments, We used SGD with a polynomial decaying learning rate, starting at 0.00025 for the segmentation model and 0.0001 for the discriminator while others such as batch size follow \cite{liu2020ocda}
Hyperparameters introduced for AH-OCDA such as $\lambda$, $\beta$, $K$, $\tau$, and the Hopfield segmentation model memory size are set to 0.001, 0.09, 3, 1, and 64, respectively.

\begin{table}[t]
\centering
\footnotesize

\caption{Experiment results on GTA5 $\to$ C-Driving benchmarks in mean IoU. Methods in the middle and lower sections list DA and OCDA methods in semantic segmentation, respectively.}
\label{tab:gta5tocdriving}
\setlength{\tabcolsep}{1pt}
\renewcommand{\arraystretch}{1.2}
\renewcommand{\tabcolsep}{1.0mm}
\begin{tabular}{lcccccc}
        \toprule
        \multirow{2}{*}{Method} & \multicolumn{3}{c}{Compound (C)} & \multicolumn{1}{c}{Open (O)} & \multicolumn{2}{c}{Total} \\
        \cmidrule(l{4pt}r{4pt}){2-4} \cmidrule(l{4pt}r{4pt}){5-5} \cmidrule(l{4pt}r{4pt}){6-7}
        & Rainy & Snowy & Cloudy & Overcast & C & C+O \\
        \midrule
        Source Only & 16.2 & 18.0 & 20.9 & 21.2 & 18.9 & 19.1 \\
        \hline
        AdaptSegNet \cite{tsai2018learning} & 20.2 & 21.2 & 23.8 & 25.1 & 22.1 & 22.5 \\
        CBST \cite{zou2018curri1} & 21.3 & 20.6 & 23.9 & 24.7 & 22.2 & 22.6 \\
        IBN-Net \cite{pan2018two} & 20.6 & 21.9 & 26.1 & 25.5 & 22.8 & 23.5 \\
        PyCDA \cite{lian2019constructing} & 21.7 & 22.3 & 25.9 & 25.4 & 23.3 & 23.8 \\
        \hline
        OCDA \cite{liu2020ocda} & 22.0 & 22.9 & 27.0 & 27.9 & 24.5 & 25.0 \\
        DHA \cite{park2020dha} & 27.0 & 26.3 & 30.7 & 32.8 & 28.5 & 29.2 \\
        MOCDA \cite{gong2021csfu} & 24.4 & 27.5 & 30.1 & 31.4 & 27.7 & 29.4 \\
        AST-OCDA \cite{kundu2022ampspectrans} & \textbf{28.2} & 27.8 & 31.6 & \textbf{34.0} & 29.2 & 30.4 \\
        \hline
        AH-OCDA (Ours)      & 28.1 & \textbf{28.1}  & \textbf{32.0}  & 33.6  & \textbf{29.7} & \textbf{31.4} \\
        \bottomrule
\end{tabular}
\end{table}

\begin{table}[t]
\centering
\small
\caption{Experiment results on GTA5 $\to$ extended open domains, Cityscape and KITTI benchmarks, reported as mean IoU. ``w/o Curr'' and ``w/o Hopf'' denotes AH-OCDA with the Hopfield segmentation model and amplitude-based curriculum learning, respectively.}
\label{tab:gtatoextend}
\setlength{\tabcolsep}{1pt}
\renewcommand{\arraystretch}{1.2}
\renewcommand{\tabcolsep}{2mm}
\begin{tabular}{lcccc}
    \toprule
        \multirow{2}{*}{Method} & \multicolumn{2}{c}{Extended Open} & \multirow{2}{*}{Total} \\
        \cmidrule(l{2pt}r{2pt}){2-3}
        & Citys. & KITTI &  \\
        \midrule
        Source only & 19.3 & 24.1 & 21.7 \\
        AdaptSegNet \cite{tsai2018learning} & 22.0 & 23.4 & 22.7 \\
        MOCDA \cite{gong2021csfu} & 31.1 & 30.9 & 31.0 \\
        AST-OCDA \cite{kundu2022ampspectrans} & \textbf{32.6} & 31.8 & 32.2 \\
        \hline
        w/o Curr & 30.2 & 32.7 & 31.5 \\
        w/o Hopf & 28.3 & 31.0 & 29.7 \\
        AH-OCDA (Ours) & \textbf{32.6} & \textbf{34.0} & \textbf{33.3} \\
    \bottomrule
\end{tabular}
\end{table}

\subsection{Baselines}
The efficacy of our framework is assessed through a comprehensive comparison with a selection of established baselines.
\textbf{Source Only} refers to a VGG segmentation model trained on the source domain.
\textbf{UDA methods} are unsupervised single-source to single-target domain adaptation frameworks.
These include AdaptSegNet \cite{tsai2018learning}, CBST \cite{zou2018curri1}, IBN-Net \cite{pan2018two}, PyCDA \cite{lian2019constructing}, CRST \cite{zou2019confidence} and AdvEnt \cite{vu2019advent}.
For \textbf{OCDA frameworks}, OCDA \cite{liu2020ocda} is the baseline framework for open compound domain adaptation.
DHA \cite{park2020dha} and MOCDA \cite{gong2021csfu} leverage clustering to decompose OCDA into multiple UDA.
AST-OCDA \cite{kundu2022ampspectrans} applies FFT-based style transfer on the feature space of the segmentation model.

\subsection{Results}

\paragraph{GTA5 as Source}
The semantic segmentation performance on GTA5 $\to$ C-Driving is reported in Table \ref{tab:gta5tocdriving}.
Comparing performance on both Compound and Compound + Open, hereinafter denoted as C and C + O respectively, AH-OCDA achieves the best mIoU.
In Table~\ref{tab:gtatoextend}, we report the semantic segmentation performance on GTA5 $\to$ extended open domains, Cityscape and KITTI.
AH-OCDA outperforms the SOTA method AST-OCDA \cite{kundu2022ampspectrans} by 1.1\%p on total mIoU, which verifies that it has better domain generalization ability.
It is because the Hopfield segmentation model helps to predict segmentation features based on source information leading the performance improvement on the extended open domains.

\begin{table}[t]
\centering
\footnotesize
\caption{Experiment results on SYNTHIA $\to$ C-Driving benchmarks in mean IoU. Methods in the middle and lower sections list DA and OCDA methods in semantic segmentation, respectively.}
\label{tab:synthiatocdriving}
\setlength{\tabcolsep}{1pt}
\renewcommand{\arraystretch}{1.2}
\renewcommand{\tabcolsep}{1.0mm}
\begin{tabular}{lcccccc}
    \toprule
        \multirow{2}{*}{Method} & \multicolumn{3}{c}{Compound (C)} & \multicolumn{1}{c}{Open (O)} & \multicolumn{2}{c}{Total} \\
        \cmidrule(l{4pt}r{4pt}){2-4} \cmidrule(l{4pt}r{4pt}){5-5} \cmidrule(l{4pt}r{4pt}){6-7}
        & Rainy & Snowy & Cloudy & Overcast & C & C+O \\
        \midrule
        Source only & 16.3 & 18.8 & 19.4 & 19.5 & 18.4 & 18.5 \\
        \hline
        CBST \cite{zou2018curri1} & 16.2 & 19.6 & 20.1 & 20.3 & 18.9 & 19.1 \\
        CRST \cite{zou2019confidence} & 16.3 & 19.9 & 20.3 & 20.5 & 19.1 & 19.3 \\
        AdaptSegNet \cite{tsai2018learning} & 17.0 & 20.5 & 21.6 & 21.6 & 20.0 & 20.2 \\
        AdvEnt \cite{vu2019advent} & 17.7 & 19.9 & 20.2 & 20.5 & 19.3 & 19.6 \\
        \hline
        DHA \cite{park2020dha} & 18.8 & 21.2 & 23.6 & 23.6 & 21.5 & 21.8 \\
        AST-OCDA \cite{kundu2022ampspectrans} & 20.3 & 22.6 & 24.9 & 25.4 & 22.6 & 23.3 \\
        \hline
        AH-OCDA (Ours) & \textbf{20.7} & \textbf{22.9} & \textbf{25.8} & \textbf{26.2} & \textbf{23.6} & \textbf{24.8} \\
    \bottomrule
\end{tabular}
\end{table}

\begin{table}[t]
\centering
\small
\caption{Experiment results on SYNTHIA $\to$ extended open domains, Cityscape and KITTI benchmarks, reported as mean IoU.}
\label{tab:synthiatoextend}
\setlength{\tabcolsep}{1pt}
\renewcommand{\arraystretch}{1.2}
\renewcommand{\tabcolsep}{2mm}
\begin{tabular}{lcccc}
    \toprule
        \multirow{2}{*}{Method} & \multicolumn{2}{c}{Extended Open} & \multirow{2}{*}{Total} \\
        \cmidrule(l{2pt}r{2pt}){2-3}
        & Citys. & KITTI &  \\
        \midrule
        Source only & 24.7 & 20.7 & 22.7 \\
        AdaptSegNet \cite{tsai2018learning} & 35.9 & 24.7 & 30.3 \\
        MOCDA \cite{gong2021csfu} & 32.2 & 34.2 & 33.2 \\
        AST-OCDA \cite{kundu2022ampspectrans} & \textbf{37.2} & 35.7 & 36.5 \\
        \hline
        AH-OCDA (Ours) & 36.8 & \textbf{37.2} & \textbf{37.0}  \\
    \bottomrule
\end{tabular}
\end{table}

\paragraph{SYNTHIA as Source}
The semantic segmentation performance on SYNTHIA $\to$ C-Driving is shown in Table \ref{tab:synthiatocdriving}.
AH-OCDA achieves the best semantic segmentation performance among all the baselines on all domains.
In comparison with AST-OCDA, AH-OCDA outperforms 1.0 \%p on C and 1.5 \%p on C + O, outperforming on all datasets, including one of the extended open dataset (Table~\ref{tab:synthiatoextend}).

\subsection{Ablation} \label{sec_ablation}

\begin{table}[t]
\centering
\footnotesize
\caption{Ablation study results by varying hyperparameter K on GTA5 $\to$ C-Driving benchmark, reported in mean IoU. Following DHA, compound domains are further divided for ``Night''.}
\label{tab:ablationk}
\setlength{\tabcolsep}{1pt}
\renewcommand{\arraystretch}{1.4}
\renewcommand{\tabcolsep}{0.8mm}
\begin{tabular}{lccccccc}
        \toprule
        \multirow{2}{*}{Method} & \multicolumn{4}{c}{Compound (C)} & \multicolumn{1}{c}{Open (O)} & \multicolumn{2}{c}{Total} \\
        \cmidrule(l{4pt}r{4pt}){2-5} \cmidrule(l{4pt}r{4pt}){6-6} \cmidrule(l{4pt}r{4pt}){7-8}
                    & Rainy & Snowy & Cloudy & Night & Overcast & C & C+O \\
        \midrule
        Source only & 23.3  & 24.0  & 28.2  & 8.1   & 30.2  & 25.7                      & 26.4 \\
        \hline
        DHA(K=2)    & 26.0  & 26.6  & 32.4  & 11.1  & 33.6  & \makecell{29.3 \\ {\color{blue}(-0.5)}} & \makecell{29.7 \\ {\color{blue}(-0.7)}} \\
        DHA(K=3)    & 26.4  & 27.5  & 33.3  & 11.8  & 34.3  & 29.8                      & 30.4 \\
        DHA(K=4)    & 25.2  & 26.4  & 32.7  & 12.1  & 33.8  & \makecell{29.1 \\ {\color{blue}(-0.7)}} & \makecell{29.5 \\ {\color{blue}(-0.9)}} \\
        DHA(K=5)    & 25.4  & 27.0  & 32.5  & 13.3  & 33.1  & \makecell{29.2 \\ {\color{blue}(-0.6)}} & \makecell{29.5 \\ {\color{blue}(-0.9)}} \\
        \hline
        Ours(K=2)   & 30.6  & 28.4  & 31.7  & 18.4  & 33.1  & \makecell{29.8 \\ {\color{blue}(-0.4)}} & \makecell{31.3 \\ {\color{blue}(-0.7)}} \\
        Ours(K=3)   & 30.1  & 28.6  & 32.4  & 18.5  & 34.0  & 30.2                      & 32.0 \\
        Ours(K=4)   & 29.9  & 28.5  & 32.5  & 19.8  & 34.1  & \makecell{30.1 \\ {\color{blue}(-0.1)}} & \makecell{32.0 \\ {\color{red}(+0.0)}} \\
        Ours(K=5)   & 29.2  & 28.4  & 31.9  & 20.0  & 33.6  & \makecell{29.7 \\ {\color{blue}(-0.5)}} & \makecell{31.5 \\ {\color{blue}(-0.5)}} \\
        \bottomrule
\end{tabular}
\end{table}

\subsubsection{Number of splits $K$ in curriculum learning} \label{sec_ablation_k}
The hyperparameter $K$ in AH-OCDA determines the number of splits applied to the compound domain to gradually expand known distributions during curriculum learning.
Similarly, DHA~\cite{park2020dha} employs a comparable hyperparameter to divide compound domains into sub-latent domains through k-means clustering.
To evaluate the robustness with $K$, we compare the hyperparameter $K$ used in AH-OCDA and DHA and report the results in Table \ref{tab:ablationk}.
In both methods, the best mIoU is achieved when the hyperparameter is set to 3.
However, when considering the robustness to $K$, a critical aspect that must be considered since the number of domains within a compound domain is unknown in advance, AH-OCDA demonstrates a much more stable performance compared to DHA.
Specifically, AH-OCDA shows no performance drop when $K$ is set to 4 and exhibits less or comparable performance drop for other values of $K$.

\subsubsection{Sub-domains visualization}

\begin{figure}[t]
    \centering
    \includegraphics[width=1.0\linewidth]{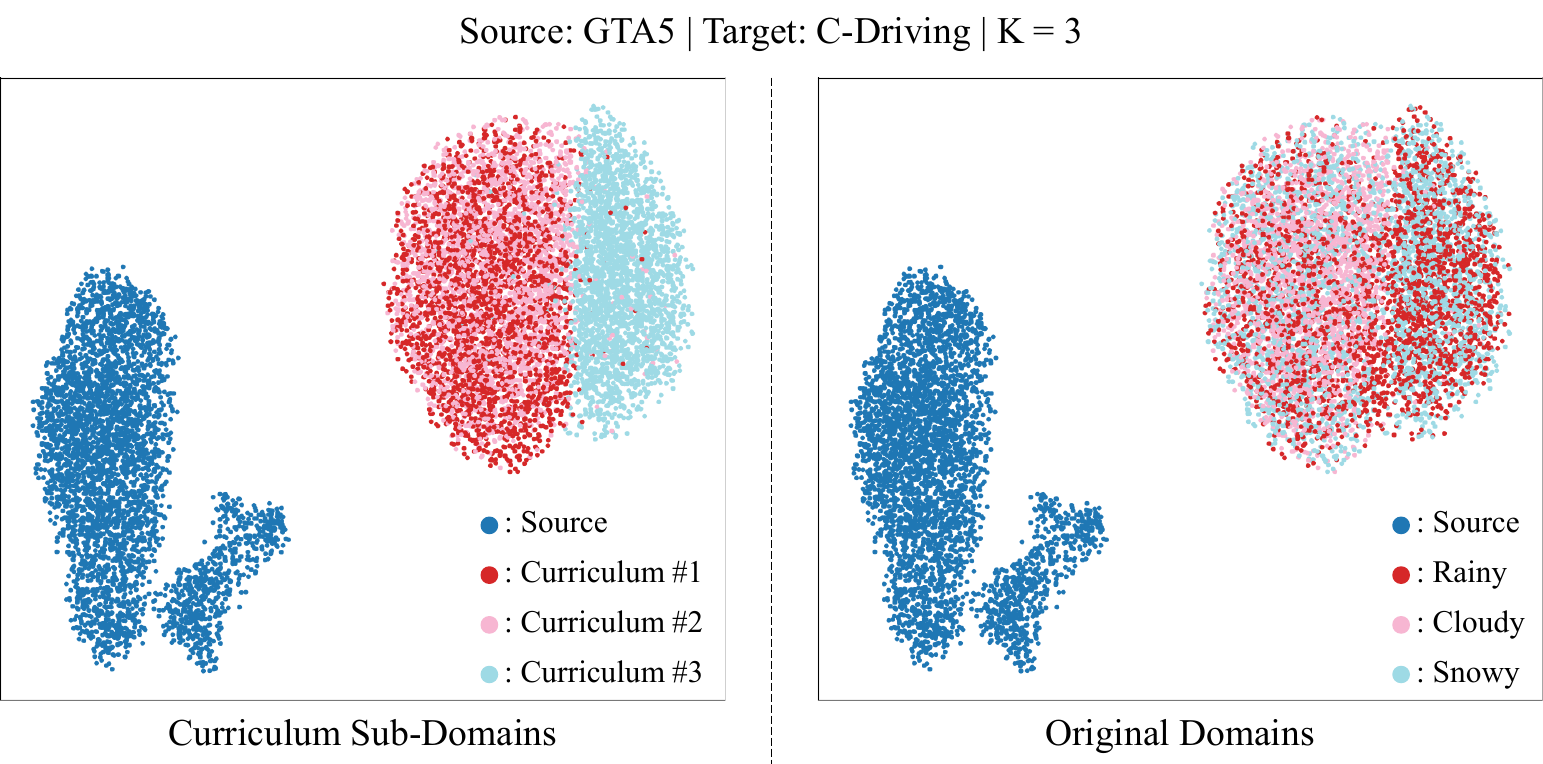}
    \caption{UMAP results of GTA5 and C-Driving when $K$ = 3.}
    \label{fig:umap}
\end{figure}

\begin{figure}[t]
    \centering
    \includegraphics[width=0.46\textwidth]{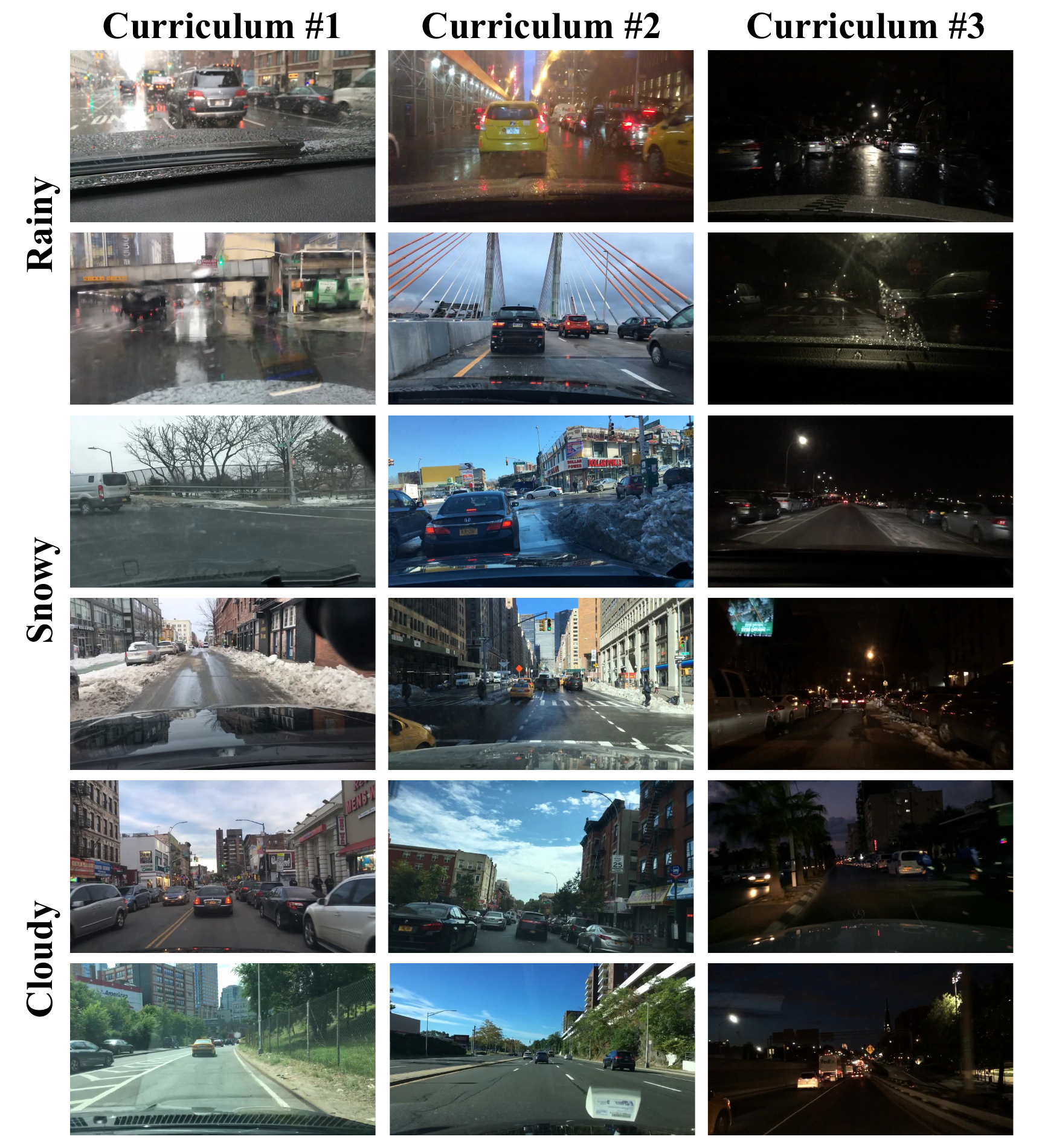}
    \caption{Sample images from each curriculum when $K$ = 3.}
    \label{fig:curriculum_samples}
\end{figure}

Figure~\ref{fig:curriculum_samples} demonstrates randomly selected images from each curriculum domain when $K$ is set to 3. 
Through amplitude-based ranking, the night or darker images are ranked later in the curriculum learning.
We also visualize the curriculum domains in the feature space by extracting features using VGG-16, the segmentation network used in all experiments.
Figure~\ref{fig:umap} validates that curriculum learning ranks the images within a compound domain as intended, ranking the images from near-source to far-source.

\subsubsection{Hopfield segmentation model settings} \label{sec_ablation_hopf}
We conduct an ablation study on the hyperparameters for the Hopfield segmentation model, memory size, and training strategy and report it in Table \ref{tab:mem}.
Experimentally, the size of memory does not have a crucial effect on segmentation performance when above 64.
As the memory size of 64 provides the best trade-off between performance and computational complexity, we used 64 for all experiments as aforementioned.
For the training strategy, we experiment with not freezing and freezing the Hopfield segmentation model when training on the compound domain.
The performance drops when the memory is not frozen as the memory gradually forgets the segmentation feature of the source domain thus we leverage the frozen model.

\begin{table}[t!]
\centering
\small
\caption{Ablation study results by varying hyperparameters of the Hopfield segmentation model on GTA5 $\to$ C-Driving benchmarks. $M_N$ represents memory size, and `w/o Freeze' and `Freeze' denote training and freezing the memory on target data, respectively.}
\renewcommand{\arraystretch}{1.2}
\renewcommand{\tabcolsep}{1.5mm}
\begin{tabular}{lcccc}
    \toprule
        {Method} & {Compound}& {Open} & {Total} \\
        \midrule
        $M_N=32$  & 28.7 & 32.0 & 30.2 \\
        $M_N=64$  & 29.5 & 33.3 & \textbf{31.2} \\
        $M_N=96$  & 29.5 & 33.2 & 31.1 \\
        \midrule
        w/o Freeze & 21.1 & 23.4 & 22.1 \\
        Freeze  & 29.5 & 33.3 & \textbf{31.2} \\
    \bottomrule
\label{tab:mem}
\end{tabular}
\end{table}

\subsubsection{Complementary components of AH-OCDA} \label{sec_ablation_k}
\begin{table}[t]
\centering
\footnotesize
\caption{Ablation study results of each component in AH-OCDA on GTA5 $\to$ C-Driving benchmarks. ``w/o Curr'' denotes AH-OCDA with Hopfield segmentation model and ``w/o Hop''  denotes AH-OCDA with amplitude-based curriculum learning.}
\label{tab:component_ab}
\setlength{\tabcolsep}{1pt}
\renewcommand{\arraystretch}{1.4}
\renewcommand{\tabcolsep}{0.8mm}
\begin{tabular}{lccccccc}
        \toprule
        \multirow{2}{*}{Method} & \multicolumn{4}{c}{Compound (C)} & \multicolumn{1}{c}{Open (O)} & \multicolumn{2}{c}{Total} \\
        \cmidrule(l{4pt}r{4pt}){2-5} \cmidrule(l{4pt}r{4pt}){6-6} \cmidrule(l{4pt}r{4pt}){7-8}
                    & Rainy & Snowy & Cloudy & Night & Overcast & C & C+O \\
        \midrule
        Source Only & 23.3  & 24.0  & 28.2  & 8.1   & 30.2      & 25.7                      & 26.4 \\
        \hline
        w/o Curr    & 26.7  & 25.3  & 29.4  & 12.2  & 30.5      & \makecell{26.9 \\ {\color{red}(+1.2)}}  & \makecell{28.5 \\ {\color{red}(+2.1)}}\\
        w/o Hopf    & 25.7  & 27.8  & 29.8  & 15.5  & 31.4      & \makecell{27.7 \\ {\color{red}(+2.0)}}  & \makecell{29.4 \\ {\color{red}(+3.0)}}\\ 
        \hline
        AH-OCDA     & \textbf{30.1}  & \textbf{28.6}  & \textbf{32.4}  & \textbf{18.5}  & \textbf{34.0}      & \makecell{\textbf{30.2} \\ {\color{red}(+4.5)}}  & \makecell{\textbf{32.0} \\ {\color{red}(+5.6)}}\\        
        \bottomrule
\end{tabular}
\end{table}

Table \ref{tab:component_ab} shows the effect of each component in AH-OCDA.
AH-OCDA outperforms `w/o Hopf' (Ours without the Hopfield segmentation model) and `w/o Curr' (Ours without amplitude-based curriculum learning).
In Table \ref{tab:component_ab}, there is a significant performance drop in the `Night' category for `w/o Curr' model aligning with the statement in Sec. \ref{met_hopfield} that the Hopfield segmentation model may not function well when adapting to a far-source target domain in the compound domain.
The amplitude-based curriculum learning complements this issue by gradually training the model from the near-source target domain so that the Hopfield segmentation model can be trained stably.

\subsubsection{Other backbone networks}
To validate that AH-OCDA is not limited to specific backbone networks, we additionally experiment using ResNet and Transformer-based backbone and compare it with unsupervised domain adaptation methods in semantic segmentation, ProDA~\cite{proda} (ResNet-101) and DAFormer~\cite{daformer} (Segformer \cite{xie2021segformer}), respectively.
Table~\ref{backbone_exp} shows the performance of VGG-16, ResNet-101, and Transformer-based architectures in the top, middle, and bottom rows, respectively.
Although the performances cannot be directly compared due to the comparison methods being proposed for another task, as aforementioned, the performance gain indirectly demonstrates that AH-OCDA is not restricted to certain backbone networks.
Notably, AH-OCDA shows better performance on a few of the target domains in the compound domain than the unsupervised domain adaptation methods, demonstrating the need for separate methods for OCDA setting.

\tabcolsep=2.5pt
\begin{table}[t!]
\centering
\small
\caption{Ablation study results on GTA5 $\to$ C-Driving benchmarks, varying the backbone network from VGG-16, ResNet-101, and Transformer-based (top to bottom). Each backbone network is compared with the well-known DA methods in semantic segmentation using the corresponding backbone network.}
\begin{tabular}{lcccccc}
    \hline
        \multirow{2}{*}{Method} & \multicolumn{3}{c}{Compound (C)} & \multicolumn{1}{c}{Open (O)} & \multicolumn{2}{c}{Total} \\
        \cmidrule(l{4pt}r{4pt}){2-4} \cmidrule(l{4pt}r{4pt}){5-5} \cmidrule(l{4pt}r{4pt}){6-7}
        & Rainy & Snowy & Cloudy & Overcast & C & C+O \\
        \midrule
        AdaptSegNet & 20.2 & 21.2 & 23.8 & 25.1 & 22.1 & 22.5 \\
        \textbf{AH-OCDA} & 28.1 & 28.1 & 32.0 & 33.6 & \textbf{29.7} & \textbf{31.4} \\
        \midrule
        ProDA & 31.61 & 30.32 & 36.29 & 34.63 & 33.36 & 33.93 \\
        \textbf{AH-OCDA} & 37.07 & 31.82 & 30.74 & 40.68 & \textbf{33.74} & \textbf{36.70} \\
        \midrule
        DAFormer & 34.04 & 31.39 & 31.14 & 34.60 & 32.10 & 33.18 \\
        \textbf{AH-OCDA} & 31.43 & 28.85 & 39.02 & 37.88 & \textbf{34.42} & \textbf{35.90} \\
        \bottomrule
\end{tabular}
\label{backbone_exp}
\end{table}

%% file: sections/5_conclusion.tex
\section{Conclusion} \label{sec_conclusion}
In this paper, we propose a novel method, AH-OCDA, considering the differences between OCDA with conventional domain adaptation and generalization settings.
Specifically, we propose two complementary components: 1) amplitude-based curriculum learning and 2) the Hopfield segmentation model.
First, by utilizing the FFT, we obtain the amplitude of each target image and measure the difference between the mean amplitude of the source domain images.
With the obtained distance, we rank them to formulate the OCDA training in a curriculum-learning manner, without any prior knowledge of domains in the compound domain.
Second, the Hopfield segmentation model learns the segmentation feature distribution on the source domain with memory and aligns features from unknown domains to the source domain.
We verify the effectiveness of the AH-OCDA with experiments on two datasets, GTA5$\to$C-Driving and SYNTHIA$\to$C-Driving.
Additionally, we present the experiment results on two novel unseen domains, Cityscape and KITTI, and achieve comparable or state-of-the-art performance.

%% file: sections/6_ack.tex
\newline
\textbf{Acknowledgements.}
This work was supported by Artificial intelligence industrial convergence cluster development project funded by the Ministry of Science and ICT(MSIT, Korea) \& Gwangju Metropolitan City.

%% file: supplementary_sections/0_dataset.tex
\section{Datasets}
\label{supple:dataset}
\paragraph{Labeled synthetic source domain}
GTA5 \cite{richter2016gta5} consists of 24,966 synthesized images of road scenes taken from a video game.
Each image sized to 1280×720, and random cropped to 1024×512 for training.
Following previous works, we selected 19 classes (total 33 classes provided by the original game engine).
SYNTHIA \cite{ros2016synthia} consists of 2224 synthetic images of virtual urban road scenes.
The images are randomly cropped to 1024×512 during training.
Following previous works, we selected 11 classes (road, sidewalk, building, wall, fence, pole, light, vegetation, sky, person, and car).
\paragraph{Unlabeled compound and open domain.}
BDD100K \cite{yu2020bdd100k} comprises 14,697 unlabeled real images of mixed domains and an additional 1430 images with labels for validation.
BDD100K includes `rainy', `snowy', and `cloudy' as compound domains, and `overcast' as as unseen domain.
For the scope of our research, other domains in the original BDD100K dataset were not employed.
\paragraph{Extended unlabeled open domain.}
Cityscapes \cite{cordts2016cityscapes} provides 2,975 real urban road scenes from various European cities for semantic segmentation task.
We utilize its validation set with 500 images for extended open evaluation.\textbf{}
KITTI \cite{geiger2013kitti} is real road scene images collected from Karlsruhe, Germany.
We utilize its validation set with 200 images for extended open evaluation.

%% file: supplementary_sections/1_hopfield_high-level.tex
\section{Intuitive Insights into the Hopfield Segmentation Model}
\begin{figure*}[t]
    \centering
    \includegraphics[width=\textwidth]{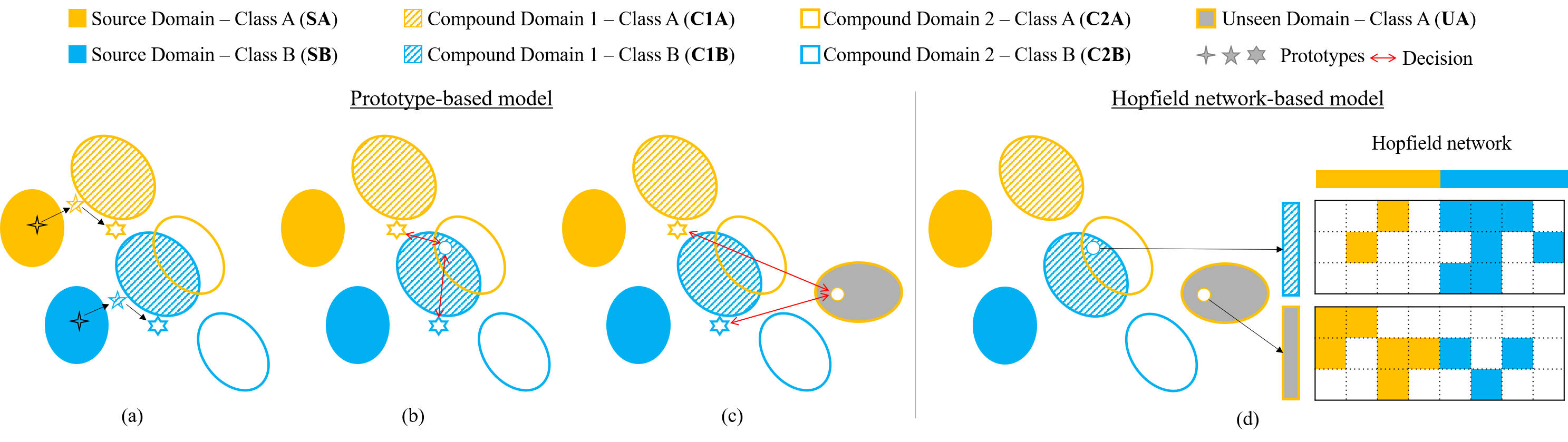}
    \vspace{-2mm}
    \caption{The high-level illustration of prototype-based model and Hopfield network-based model on OCDA setting.}
    \label{fig:rebut_hopfield}
    \vspace{-3mm}
\end{figure*}

Figure \ref{fig:rebut_hopfield} offers high-level insights into why the Hopfield segmentation model in AH-OCDA achieves strong performance and highlights the effectiveness of its approach in addressing the OCDA problem.
We compare the Hopfield segmentation model with the prototype-based DA method in the OCDA setting especially focusing on how would each handle compound domains and obtain representations.

Prototype-based models use explicit feature vectors as class representations by updating the prototype through the training of compound domains (Fig.\ref{fig:rebut_hopfield}(a)) thus it may incorrectly predict when C1B is close to the prototype of class A (Fig.\ref{fig:rebut_hopfield}(b)).
A similar result occurs when UA is close to the prototype of class B (Fig.\ref{fig:rebut_hopfield}(c)).
In contrast, our Hopfield segmentation model implicitly stores both intra-class and inter-class information in its parameters, enabling accurate predictions even for distorted features from compound or unseen domains (Fig.\ref{fig:rebut_hopfield}(d)).
Furthermore, this clearly distinguishes from segmentation models that utilize the conventional Hopefield network as an auxiliary optimization module.
In addition, it supports the findings of the ablation study presented in Section 4.5.3, which demonstrates that freezing the Hopfield segmentation model during training on the compound domain yields better performance.

%% file: supplementary_sections/2_qualitative_result.tex
\section{Qualitative Result}

The qualitative results of GTA5 $\to$ C-Driving are presented in Fig. \ref{fig:sup_cloudy} $\sim$ Fig. \ref{fig:sup_overcast}.
In addition to the AH-OCDA, we present the qualitative results of each component of AH-OCDA.
`w/o Curr' and `w/o Hopf' denote AH-OCDA without curriculum learning and AH-OCDA without the Hopfield network, respectively.
While AH-OCDA shows phenomenal segmentation results, there are some consistent tendencies corresponding to each component of AH-OCDA.

Specifically, it is possible to infer the significance of amplitude-based curriculum learning through the comparative analysis of `w/o Hopf' and `Source only'.
As discussed in the main manuscript, amplitude-based curriculum learning gradually trains the segmentation model to achieve proficient performance even on the compound domain far from the source domain.
Upon comparing `w/o Hopf' with `Source only', it is evident that amplitude-based curriculum learning substantially improved segmentation results.
Especially for the `Night' domain, which is far from the source domain, `Source Only' shows poor performance where it is difficult to distinguish the car shape.
In contrast, `w/o Hopf' shows improved segmentation results, enabling the differentiation of the car's pixels from the road.
These results support that amplitude-based curriculum learning led to the successful adaptation of the segmentation model to compound domains, including a domain that is close to the source domain to the farthest domain.
Moreover, the Hopfield segmentation model maps the feature distribution of the target image to the source domain.
The `w/o Curr' results show stable segmentation across all domains, compared to `Source Only' and `w/o Hopf'.
Specifically, segmentation results for `Source Only' and `w/o Hopf' exhibit unstable predictions often showing a mosaic pattern.
The different patterns in segmentation results support that the Hopfield segmentation model retrieves the pattern of features in the source domain for the consistent prediction of one object.
AH-OCDA produces precise and stable segmentation results in both compound and open domains, as the amplitude-based curriculum learning and the Hopfield segmentation model complement each other, leveraging their respective strengths.

Lastly, we acknowledge the efforts of those who contributed to the BDD100K dataset and inevitable errors in labeling even with meticulous efforts.
Through examples of the qualitative results, we draw attention to the presence of mislabeled samples within the BDD100K dataset to inform future researchers in the field.
As of what we have found, in Fig. \ref{fig:sup_snowy}, the road is labeled as the sky, and in Fig. \ref{fig:sup_night}, most of the pixels except for the car are labeled as the road when it's not.
Nonetheless, we utilized the original labels provided from the dataset for a fair comparison with prior works.

%% file: supplementary_sections/3_limitations.tex
\section{Limitations}

As demonstrated in Table 5 of the main manuscript, AH-OCDA exhibits robustness to variations in the number of $K$ when compared to the previous method.
However, when there is an imbalance in image distribution within a compound domain---for example, when images are predominantly distributed at the beginning of the curriculum--- AH-OCDA could experience performance degradation due to dividing the sorted compound domain images by $K$ splits.
In such a situation, we may alternatively define a threshold value rather than the number of $K$.
When the distance between two adjacent images in the sorted compound domain images exceeds the defined threshold, we can consider that they have distant domain information.
This approach may even be beneficial when there is a huge domain gap between the domains in the compound domain.
Nevertheless, we opted not to implement thresholding in AH-OCDA as thresholding the distance value requires more precise consideration and is considerably more sensitive compared to the use of the number of $K$.
In our future research, we aim to remove the hyperparameter $K$ to enhance generalizability.

%% file: main.bbl
\begin{thebibliography}{10}
\providecommand{\url}[1]{\texttt{#1}}
\providecommand{\urlprefix}{URL }
\providecommand{\doi}[1]{https://doi.org/#1}

\bibitem{blanchard2021dg3}
Blanchard, G., Deshmukh, A.A., Dogan, {\"U}., Lee, G., Scott, C.: Domain generalization by marginal transfer learning. The Journal of Machine Learning Research  \textbf{22}(1),  46--100 (2021)

\bibitem{dg1}
Choi, J., Seong, H.S., Park, S., Heo, J.P.: Tcx: Texture and channel swappings for domain generalization. Pattern Recognition Letters  \textbf{175},  74--80 (2023)

\bibitem{cordts2016cityscapes}
Cordts, M., Omran, M., Ramos, S., Rehfeld, T., Enzweiler, M., Benenson, R., Franke, U., Roth, S., Schiele, B.: The cityscapes dataset for semantic urban scene understanding. In: Proceedings of the IEEE conference on computer vision and pattern recognition. pp. 3213--3223 (2016)

\bibitem{dou2019dg1}
Dou, Q., Coelho~de Castro, D., Kamnitsas, K., Glocker, B.: Domain generalization via model-agnostic learning of semantic features. Advances in neural information processing systems  \textbf{32} (2019)

\bibitem{geiger2013kitti}
Geiger, A., Lenz, P., Stiller, C., Urtasun, R.: Vision meets robotics: The kitti dataset. The International Journal of Robotics Research  \textbf{32}(11),  1231--1237 (2013)

\bibitem{gholami2020DAmulti2}
Gholami, B., Sahu, P., Rudovic, O., Bousmalis, K., Pavlovic, V.: Unsupervised multi-target domain adaptation: An information theoretic approach. IEEE Transactions on Image Processing  \textbf{29},  3993--4002 (2020)

\bibitem{gong2013DAmulti1}
Gong, B., Grauman, K., Sha, F.: Reshaping visual datasets for domain adaptation. Advances in Neural Information Processing Systems  \textbf{26} (2013)

\bibitem{gong2021csfu}
Gong, R., Chen, Y., Paudel, D.P., Li, Y., Chhatkuli, A., Li, W., Dai, D., Van~Gool, L.: Cluster, split, fuse, and update: Meta-learning for open compound domain adaptive semantic segmentation. In: Proceedings of the IEEE/CVF Conference on computer vision and pattern recognition. pp. 8344--8354 (2021)

\bibitem{hopfield2007hopfield}
Hopfield, J.J.: Hopfield network. Scholarpedia  \textbf{2}(5), ~1977 (2007)

\bibitem{daformer}
Hoyer, L., Dai, D., Van~Gool, L.: Daformer: Improving network architectures and training strategies for domain-adaptive semantic segmentation. In: Proceedings of the IEEE/CVF conference on computer vision and pattern recognition. pp. 9924--9935 (2022)

\bibitem{huang2021fft2}
Huang, J., Guan, D., Xiao, A., Lu, S.: Rda: Robust domain adaptation via fourier adversarial attacking. In: Proceedings of the IEEE/CVF international conference on computer vision. pp. 8988--8999 (2021)

\bibitem{huang2020dg2}
Huang, Z., Wang, H., Xing, E.P., Huang, D.: Self-challenging improves cross-domain generalization. In: Computer Vision--ECCV 2020: 16th European Conference, Glasgow, UK, August 23--28, 2020, Proceedings, Part II 16. pp. 124--140. Springer (2020)

\bibitem{kundu2022ampspectrans}
Kundu, J.N., Kulkarni, A.R., Bhambri, S., Jampani, V., Radhakrishnan, V.B.: Amplitude spectrum transformation for open compound domain adaptive semantic segmentation. In: Proceedings of the AAAI Conference on Artificial Intelligence. vol.~36, pp. 1220--1227 (2022)

\bibitem{lee2023lightweight}
Lee, D.J., Lee, J.Y., Shon, H., Yi, E., Park, Y.H., Cho, S.S., Kim, J.: Lightweight monocular depth estimation via token-sharing transformer. In: 2023 IEEE International Conference on Robotics and Automation (ICRA). pp. 4895--4901. IEEE (2023)

\bibitem{singlesourcedg2}
Li, L., Gao, K., Cao, J., Huang, Z., Weng, Y., Mi, X., Yu, Z., Li, X., Xia, B.: Progressive domain expansion network for single domain generalization. In: Proceedings of the IEEE/CVF Conference on Computer Vision and Pattern Recognition. pp. 224--233 (2021)

\bibitem{li2023fftstyle}
Li, T.B., Su, Y.T., Song, D., Li, W.H., Wei, Z.Q., Liu, A.A.: Progressive fourier adversarial domain adaptation for object classification and retrieval. IEEE Transactions on Multimedia  (2023)

\bibitem{lian2019constructing}
Lian, Q., Lv, F., Duan, L., Gong, B.: Constructing self-motivated pyramid curriculums for cross-domain semantic segmentation: A non-adversarial approach. In: Proceedings of the IEEE/CVF International Conference on Computer Vision. pp. 6758--6767 (2019)

\bibitem{liu2020ocda}
Liu, Z., Miao, Z., Pan, X., Zhan, X., Lin, D., Yu, S.X., Gong, B.: Open compound domain adaptation. In: Proceedings of the IEEE/CVF Conference on Computer Vision and Pattern Recognition. pp. 12406--12415 (2020)

\bibitem{uhn}
Millidge, B., Salvatori, T., Song, Y., Lukasiewicz, T., Bogacz, R.: Universal hopfield networks: A general framework for single-shot associative memory models. In: International Conference on Machine Learning. pp. 15561--15583. PMLR (2022)

\bibitem{pan2018two}
Pan, X., Luo, P., Shi, J., Tang, X.: Two at once: Enhancing learning and generalization capacities via ibn-net. In: Proceedings of the European Conference on Computer Vision (ECCV). pp. 464--479 (2018)

\bibitem{park2020dha}
Park, K., Woo, S., Shin, I., Kweon, I.S.: Discover, hallucinate, and adapt: Open compound domain adaptation for semantic segmentation. Advances in Neural Information Processing Systems  \textbf{33},  10869--10880 (2020)

\bibitem{peng2019DAsingle3}
Peng, X., Bai, Q., Xia, X., Huang, Z., Saenko, K., Wang, B.: Moment matching for multi-source domain adaptation. In: Proceedings of the IEEE/CVF international conference on computer vision. pp. 1406--1415 (2019)

\bibitem{multisourceda2}
Peng, X., Bai, Q., Xia, X., Huang, Z., Saenko, K., Wang, B.: Moment matching for multi-source domain adaptation. In: Proceedings of the IEEE/CVF international conference on computer vision. pp. 1406--1415 (2019)

\bibitem{singlesourcedg1}
Qiao, F., Zhao, L., Peng, X.: Learning to learn single domain generalization. In: Proceedings of the IEEE/CVF conference on computer vision and pattern recognition. pp. 12556--12565 (2020)

\bibitem{hopfieldall}
Ramsauer, H., Sch{\"a}fl, B., Lehner, J., Seidl, P., Widrich, M., Adler, T., Gruber, L., Holzleitner, M., Pavlovi{\'c}, M., Sandve, G.K., et~al.: Hopfield networks is all you need. arXiv preprint arXiv:2008.02217  (2020)

\bibitem{richter2016gta5}
Richter, S.R., Vineet, V., Roth, S., Koltun, V.: Playing for data: Ground truth from computer games. In: Computer Vision--ECCV 2016: 14th European Conference, Amsterdam, The Netherlands, October 11-14, 2016, Proceedings, Part II 14. pp. 102--118. Springer (2016)

\bibitem{ros2016synthia}
Ros, G., Sellart, L., Materzynska, J., Vazquez, D., Lopez, A.M.: The synthia dataset: A large collection of synthetic images for semantic segmentation of urban scenes. In: Proceedings of the IEEE conference on computer vision and pattern recognition. pp. 3234--3243 (2016)

\bibitem{saenko2010DAsingle1}
Saenko, K., Kulis, B., Fritz, M., Darrell, T.: Adapting visual category models to new domains. In: Computer Vision--ECCV 2010: 11th European Conference on Computer Vision, Heraklion, Crete, Greece, September 5-11, 2010, Proceedings, Part IV 11. pp. 213--226. Springer (2010)

\bibitem{da_styletransfer2}
Term{\"o}hlen, J.A., Klingner, M., Brettin, L.J., Schmidt, N.M., Fingscheidt, T.: Continual unsupervised domain adaptation for semantic segmentation by online frequency domain style transfer. In: 2021 IEEE International Intelligent Transportation Systems Conference (ITSC). pp. 2881--2888. IEEE (2021)

\bibitem{tsai2018learning}
Tsai, Y.H., Hung, W.C., Schulter, S., Sohn, K., Yang, M.H., Chandraker, M.: Learning to adapt structured output space for semantic segmentation. In: Proceedings of the IEEE conference on computer vision and pattern recognition. pp. 7472--7481 (2018)

\bibitem{venkateswara2017DAsingle2}
Venkateswara, H., Eusebio, J., Chakraborty, S., Panchanathan, S.: Deep hashing network for unsupervised domain adaptation. In: Proceedings of the IEEE conference on computer vision and pattern recognition. pp. 5018--5027 (2017)

\bibitem{vu2019advent}
Vu, T.H., Jain, H., Bucher, M., Cord, M., P{\'e}rez, P.: Advent: Adversarial entropy minimization for domain adaptation in semantic segmentation. In: Proceedings of the IEEE/CVF conference on computer vision and pattern recognition. pp. 2517--2526 (2019)

\bibitem{da_styletransfer3}
Wang, A., Islam, M., Xu, M., Ren, H.: Curriculum-based augmented fourier domain adaptation for robust medical image segmentation. IEEE Transactions on Automation Science and Engineering  (2023)

\bibitem{xie2021segformer}
Xie, E., Wang, W., Yu, Z., Anandkumar, A., Alvarez, J.M., Luo, P.: Segformer: Simple and efficient design for semantic segmentation with transformers. Advances in neural information processing systems  \textbf{34},  12077--12090 (2021)

\bibitem{fftdg}
Xu, Q., Zhang, R., Zhang, Y., Wang, Y., Tian, Q.: A fourier-based framework for domain generalization. In: Proceedings of the IEEE/CVF conference on computer vision and pattern recognition. pp. 14383--14392 (2021)

\bibitem{multisourceda1}
Xu, R., Chen, Z., Zuo, W., Yan, J., Lin, L.: Deep cocktail network: Multi-source unsupervised domain adaptation with category shift. In: Proceedings of the IEEE conference on computer vision and pattern recognition. pp. 3964--3973 (2018)

\bibitem{da_styletransfer1}
Yang, Y., Lao, D., Sundaramoorthi, G., Soatto, S.: Phase consistent ecological domain adaptation. In: Proceedings of the IEEE/CVF conference on computer vision and pattern recognition. pp. 9011--9020 (2020)

\bibitem{yang2020fda}
Yang, Y., Soatto, S.: Fda: Fourier domain adaptation for semantic segmentation. In: Proceedings of the IEEE/CVF conference on computer vision and pattern recognition. pp. 4085--4095 (2020)

\bibitem{yu2020bdd100k}
Yu, F., Chen, H., Wang, X., Xian, W., Chen, Y., Liu, F., Madhavan, V., Darrell, T.: Bdd100k: A diverse driving dataset for heterogeneous multitask learning. In: Proceedings of the IEEE/CVF conference on computer vision and pattern recognition. pp. 2636--2645 (2020)

\bibitem{yu2018multi3}
Yu, H., Hu, M., Chen, S.: Multi-target unsupervised domain adaptation without exactly shared categories. arXiv preprint arXiv:1809.00852  (2018)

\bibitem{proda}
Zhang, P., Zhang, B., Zhang, T., Chen, D., Wang, Y., Wen, F.: Prototypical pseudo label denoising and target structure learning for domain adaptive semantic segmentation. In: Proceedings of the IEEE/CVF conference on computer vision and pattern recognition. pp. 12414--12424 (2021)

\bibitem{zhong2023bifreqalign}
Zhong, X., Li, W., Liao, L., Xiao, J., Liu, W., Huang, W., Wang, Z.: Bat: Bi-alignment based on transformation in multi-target domain adaptation for semantic segmentation. In: ICASSP 2023-2023 IEEE International Conference on Acoustics, Speech and Signal Processing (ICASSP). pp.~1--5. IEEE (2023)

\bibitem{zou2018curri1}
Zou, Y., Yu, Z., Kumar, B., Wang, J.: Unsupervised domain adaptation for semantic segmentation via class-balanced self-training. In: Proceedings of the European conference on computer vision (ECCV). pp. 289--305 (2018)

\bibitem{zou2019confidence}
Zou, Y., Yu, Z., Liu, X., Kumar, B., Wang, J.: Confidence regularized self-training. In: Proceedings of the IEEE/CVF international conference on computer vision. pp. 5982--5991 (2019)

\end{thebibliography}
